\documentclass{article} 
\usepackage{iclr2019_conference,times}

\usepackage{amsmath,amsfonts,bm}

\def\eqref#1{equation~\ref{#1}}

\def\1{\bm{1}}

\DeclareMathAlphabet{\mathsfit}{\encodingdefault}{\sfdefault}{m}{sl}
\SetMathAlphabet{\mathsfit}{bold}{\encodingdefault}{\sfdefault}{bx}{n}

\usepackage[final]{graphicx}
\usepackage{float}
\usepackage{hyperref}
\usepackage{rotating}
\usepackage{url}
\usepackage{algorithm}
\usepackage[noend]{algpseudocode}
\usepackage{graphicx,amsmath,amsfonts}
\usepackage{tikz}
\usepackage{subfigure}
\usepackage{booktabs}
\usetikzlibrary{arrows, positioning, automata}
\usepackage{wrapfig}

\title{Unsupervised Hyperalignment for Multilingual Word Embeddings}
\author{Jean Alaux$^{1,2}$ \& Edouard Grave$^1$ \& Marco Cuturi$^{2,3}$ \& Armand Joulin$^1$
\\
$^1$Facebook AI Research
\\
$^2$CREST ENSAE, Institut Polytechnique de Paris
\\
$^3$Google Brain}

\iclrfinalcopy
\begin{document}

\maketitle

\begin{abstract}
We consider the problem of aligning continuous word representations, learned in multiple languages, to a common space.
It was recently shown that, in the case of two languages, it is possible to learn such a mapping without supervision.
This paper extends this line of work to the problem of aligning multiple languages to a common space.
A solution is to independently map all languages to a pivot language.
Unfortunately, this degrades the quality of indirect word translation.
We thus	propose	a novel formulation that ensures composable mappings, leading to better alignments.
We evaluate our method by jointly aligning word vectors in eleven languages, showing consistent improvement with indirect mappings
while maintaining competitive performance on direct word translation.
\end{abstract}
\maketitle
\section{Introduction}
Pre-trained continuous representations of words are standard building blocks of many natural language processing and machine learning systems~\citep{mikolov2013distributed}.
Word vectors are designed to summarize and quantify semantic nuances through a few hundred coordinates.
Such representations are typically used in downstream tasks to improve generalization when the amount of data is scarce~\citep{collobert2011natural}.
The distributional information used to learn these word vectors derives from statistical properties of word co-occurrence found in large corpora~\citep{deerwester1990indexing}.
Such corpora are, by design, monolingual~\citep{mikolov2013distributed,FastText}, resulting in the independent learning of word embeddings for each language.

A limitation of these monolingual embeddings is that it is impossible to compare words across languages.
It is thus natural to try to combine all these word representations into a common multilingual space, 
where every language could be mapped.
\citet{Mikolov13} observed that word vectors learned on different languages share a similar structure.
More precisely, two sets of pre-trained vectors in different languages can be aligned to some extent: 
a linear mapping between the two sets of embeddings is enough to produce decent word translations.
Recently, there has been an increasing interest in mapping these pre-trained vectors in a common space~\citep{Xing15,artetxe2017learning}, resulting in many publicly available embeddings in many languages mapped into a single common vector space~\citep{smith2017offline,conneau2017word,joulin2018loss}.
The quality of these multilingual embeddings can be tested by composing mappings between languages and looking at the resulting translations.
As an example, learning a direct mapping between Italian and Portuguese leads to a word translation accuracy of $78.1\%$ with a nearest neighbor (NN) criterion,
while composing the mapping between Italian and English and Portuguese and English leads to a word translation accuracy of $70.7\%$ only.
Practically speaking, it is not surprising to see such a degradation since these bilingual alignments are trained separately, without enforcing transitivity.

In this paper, we propose a novel approach to align multiple languages
simultaneously in a common space in a way that enforces transitive translations.
Our method relies on constraining word translations to be coherent between languages when mapped to the common space.
\citet{nakashole2017knowledge} has recently shown that similar constraints over a well chosen triplet
of languages improve supervised bilingual alignment.
Our work extends their conclusions to the unsupervised case.
We show that our approach achieves competitive performance while enforcing composition.

\begin{figure}[t]
\begin{minipage}[l]{.6\linewidth}
\centering
\includegraphics[width = 3.5in]{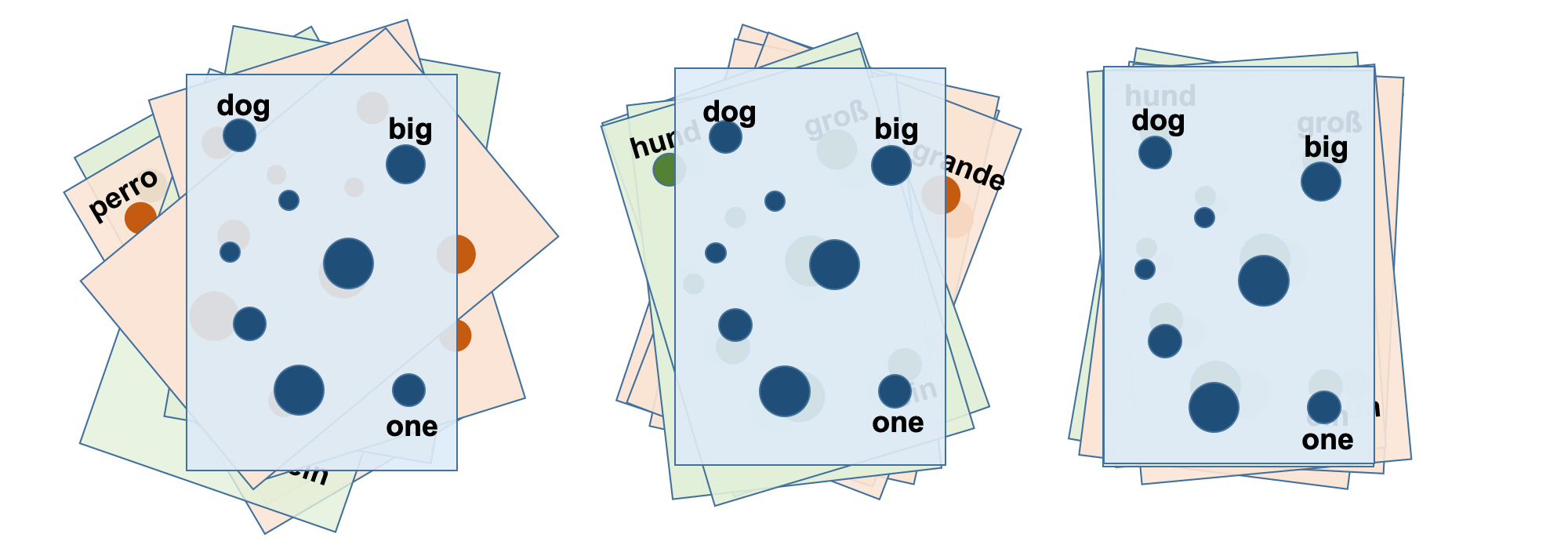}
\end{minipage}
\begin{minipage}[r]{.38\linewidth}
\caption{Left: we do not impose constraint for language pairs not involving English,
giving poor alignments for languages different than English.
We can add constraints between languages belonging to a same family (middle) or between all pairs (right), leading to better alignments.}
\end{minipage}
\end{figure}


\section{Preliminaries on bilingual alignment}

In this section, we provide a brief overview of bilingual alignment methods to learn a mapping between two sets of embeddings, and discuss their limits when used in multilingual settings.

\subsection{Supervised bilingual alignment}

\citet{Mikolov13} formulate the problem of word vector alignment as a quadratic problem.
Given two sets of $n$ word vectors stacked in two $n \times d$ matrices $\mathbf{X}$ and $\mathbf{Y}$ 
and a $n\times n$ assignment matrix $\mathbf{P} \in \{0, 1\}^{n \times n}$ built using a bilingual lexicon, 
the mapping matrix $\mathbf{Q} \in \mathbb{R}^{d \times d}$ is the solution of the least-square problem:
$\min\limits_{\mathbf{Q} \in \mathbb{\mathbb{R}}^{d\times d}}\|  \mathbf{XQ} - \mathbf{PY} \|_2^2,$
which admits a closed form solution. 
Restraining $\mathbf{Q}$ to the set of orthogonal matrices $\mathcal{O}_d$, improves the alignments~\citep{Xing15}. 
The resulting problem, known as Orthogonal Procrustes, still admits a closed form solution through a singular value decomposition~\citep{schönemann66}.

\paragraph{Alternative loss function.}
The $\ell_2$ loss is intrinsically associated with the nearest neighbor (NN) criterion. 
This criterion suffers from the existence of ``hubs'', which are data points that are nearest neighbors to many other data points~\citep{dinu2014improving}.
Alternative criterions have been suggested, such as the inverted softmax~\citep{smith2017offline} and CSLS~\citep{conneau2017word}.
Recently,~\citet{joulin2018loss} has shown that directly minimizing a loss inspired by the CSLS criterion significantly improve the quality of the retrieved word translations. 
Their loss function, called RCSLS, is defined as:
\begin{equation}
  \textsc{RCSLS}(\mathbf{x}, \mathbf{y}) =
  - 2 \mathbf{x}^\top \mathbf{y}
  + \frac{1}{k} \sum_{\mathbf{y} \in \mathcal{N}_Y (\mathbf{x})} \mathbf{x}^\top \mathbf{y} + \frac{1}{k} \sum_{\mathbf{x} \in \mathcal{N}_X (\mathbf{y})} \mathbf{x}^\top \mathbf{y}.
\label{eq:rcsls}
\end{equation}
This loss is a tight convex relaxation of the CSLS cristerion for normalized word vectors, and can be efficiently minimized with a subgradient method.

\subsection{Unsupervised bilingual Alignment: Wasserstein-Procrustes} 
\label{sec:WP}
In the setting of unsupervised bilingual alignment, the assignment matrix $\mathbf{P}$ is unknown and must be learned jointly with the mapping $\mathbf{Q}$.
An assignment matrix represents a one-to-one correspondence between the two sets of words, i.e., is a bi-stochastic matrix with binary entries.
The set of assignment matrices $\mathcal{P}_n$ is thus defined as:
$$\mathcal{P}_n= \mathcal{B}_n \cap \{0,1\}^{n\times n}, \text{ where } \mathcal{B}_n = \left\{  \mathbf{P} \in \mathbb{R}_+^{n\times n}, \mathbf{P}\1_\mathrm{n}=\1_\mathrm{n}, \  \mathbf{P}^\top \1_\mathrm{n}=\1_\mathrm{n} \right\}.$$
The resulting approach, called Wasserstein-Procrustes \citep{Zhang17b,grave18}, jointly learns both matrices by solving the following problem:
\begin{equation}\label{eq:wass-proc}
\min\limits_{\mathbf{Q} \in \mathcal{O}_d}  \min\limits_{\mathbf{P} \in \mathcal{P}_n}   \|\mathbf{XQ} - \mathbf{PY} \|_2^2.
\end{equation}
This problem is not convex since neither of the sets $\mathcal{P}_n$ and $\mathcal{O}_d$ are convex. 
Minimizing over each variable separately leads, however, to well understood optimization problems: 
when $\mathbf{P}$ is fixed, minimizing over $\mathbf{Q}$ involves solving the orthogonal Procrustes problem. 
When $\mathbf{Q}$ is fixed, an optimal permutation matrix $\mathbf{P}$ can be obtained with the Hungarian algorithm.  
A simple heuristic to address~Eq.(\ref{eq:wass-proc}) is thus to use an alternate optimization. 
Both algorithms have a cubic complexity but on different quantities: Procrustes involves the dimension of the vectors, i.e., $O(d^3)$ (with $d=300$), 
whereas the Hungarian algorithm involves the size of the sets, i.e., $O(n^3)$ (with $n=20$k--$200$k).
Directly applying the Hungarian algorithm is computationally prohibitive, but efficient alternatives exist.
\citet{cuturi13} shows that regularizing this problem with a negative entropy leads to a Sinkhorn algorithm, with a complexity of $O(n^2)$ up to logarithmic factors~\citep{altschuler2017near}. 

As for many non-convex problems, a good initial guess helps converge to better local minima. 
~\citet{grave18} compute an initial $\mathbf{P}$ with a convex relaxation of the quadratic assignment problem. 
We found, however, that the entropic regularization of the Gromov-Wasserstein (GW) problem~\citep{memoli2011gromov} worked well in practice and was significantly faster~\citep{solomon2016entropic,peyre2016gromov}: 
\begin{equation*}\label{eq:gromov}
\min\limits_{\mathbf{P} \in\mathcal{P}_n }\sum_{i,j, i', j'}\left(\|\mathbf{x}_i - \mathbf{x}_{i'}  \|_2- \|\mathbf{y}_{j} - \mathbf{y}_{j'} \|_2 \right)^2P(i, j) P(i', j') + \epsilon \sum_{i,j} P(i, j)\log P(i, j).
\end{equation*}
The case $\varepsilon=0$ corresponds to~\citeauthor{memoli2011gromov}'s initial proposal. 
Optimizing the regularized version ($\epsilon>0$) leads to a local minimum $\hat{\mathbf{P}}$ that can be used as an initialization to solve Eq.~(\ref{eq:wass-proc}). Note that a similar formulation was recently used in the same context by~\citet{alvarez2018gromov}.


\section{Composable Multilingual Alignments}

In this section, we propose an unsupervised approach to jointly align $N$ sets of vectors to a unique common space while preserving the quality of word translation between every pair of languages.

\begin{figure}[t]
\begin{minipage}[l]{.47\linewidth}
\centering
\includegraphics[width = 1.2in]{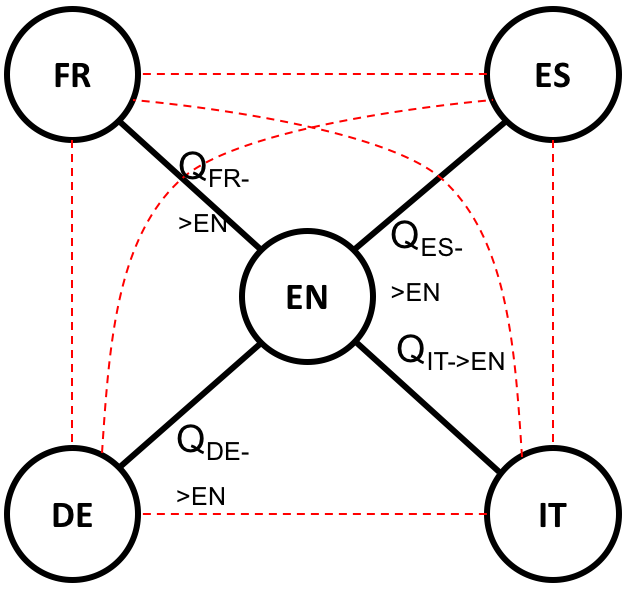}
\end{minipage}
\begin{minipage}[r]{.47\linewidth}
\caption{
  Plain black lines indicate pairs of languages for which a translation matrix is learned.
  Dashed red lines indicate pairs used in the loss functions.
  Languages are aligned onto a common pivot language, i.e., English.
}
\end{minipage}
\end{figure}

\subsection{Multilingual alignment to a common space}

Given $N$ sets of word vectors, we are interested in aligning these to a common target space $\mathcal{T}$.
For simplicity, we assume that this target space coincide with one of the word vector set.
The language associated with this vector set is denoted as ``pivot'' and is indexed by $i=0$.
A typical choice for the pivot, used in publicly available aligned vectors, is English~\citep{smith2017offline,conneau2017word,joulin2018loss}.
Aligning multiple languages to a common space consists in learning a mapping~$\mathbf{Q}_i$ for each language $i$, such that its vectors are aligned with the vectors of the pivot language up to a permutation matrix $\mathbf{P}_{i}$:
\begin{eqnarray}
\min\limits_{\mathbf{Q}_{i}\in \mathcal{O}_d,~\mathbf{P}_{i} \in \mathcal{P}_n}
  & \sum \limits_{i} \ell(\mathbf{X}_i \mathbf{Q}_{i}, \mathbf{P}_{i} \mathbf{X}_0),
\label{eq:unc}
\end{eqnarray}
This objective function decomposes over each language and does not guarantee good indirect word translation between pairs of languages that do not include the pivot.
A solution would be to  directly enforce compositionality by adding constraints on the mappings.
However, this would require to introduce mappings between all pairs of languages, leading to the estimation of $O(N^2)$ mappings simultaneously.
Instead we leverage the fact that all the vector sets are mapped to a common space to enforce good alignments within this space.
With the convention that $\mathbf{Q}_0$ is equal to the identity, this leads to the following problem:
\begin{eqnarray}
\min\limits_{\mathbf{Q}_{i}\in \mathcal{O}_d,~\mathbf{P}_{ij} \in \mathcal{P}_n}
  & \sum \limits_{i,j} \alpha_{ij} \ell(\mathbf{X}_i \mathbf{Q}_{i}, \mathbf{P}_{ij} \mathbf{X}_j \mathbf{Q}_{j}),
\label{eq:simple}
\end{eqnarray}
where $\alpha_{ij}>0$ weights the importance of the alignment quality between the languages $i$ and $j$.
This formulation does not introduce any unnecessary mapping.
It constrains all pairs of vector sets to be well aligned, instead of directly constraining the mappings.
Constraining the mappings would encourage coherence over the entire space, while we are interested in well aligned data;
that is coherent mapping within the span of the word vectors.
Our approach takes its inspiration from the hyperalignment
of multiple graphs~\citep{goodall91}.
We refer to our approach as Unsupervised Multilingual Hyperalignment (UMH).

\paragraph{Choice of weights.}
The weights on the alignments can be chosen to reflect prior knowledge about the relations between languages.
For example, these weights can be a function of a rough language similarity measure extracted from the initial alignment.
It is not clear however how the weights should depend on these similarities:
constraining similar languages with higher weights may be unnecessary, since they are already well aligned.
On the other hand, increasing weights of distant languages may lead to a problem where alignments are harder to learn.
In practice, we find that simple weighting schemes work best with little assumptions.
For instance, uniform weights work well but tends to degrade the performance of translations to and from the pivot if the number of languages is large.
Instead, we choose to use larger weights for direct alignments to the pivot, to insure that our multilingual objective does not degrade
bilingual alignments to the pivot.
In practice, $\alpha_{ij}$ is set to $N$ if $i$ or $j$ is equal to $0$, and $1$ otherwise.

\paragraph{Choice of loss function.}
We consider the RCSLS loss of \citet{joulin2018loss}.
This loss is computationally expensive and wasteful to minimize it from scratch.
We thus optimize a $\ell_2$ for the first couple of epochs before switching to the RCSLS loss.
This two-step procedure shares some similarities with the use of a refinement step after learning a first rough alignment~\citep{artetxe2017learning,conneau2017word}.
The $\ell_2$ loss between two sets of normalized vectors is equivalent to a linear function, up to an additive constant~$C$:
\begin{eqnarray}
  \ell_2(\mathbf{Q X}, \mathbf{P}\mathbf{Y})=- 2 \text{tr } (\mathbf{Q}^T \mathbf{X}^T \mathbf{P} \mathbf{Y}) + C
\label{eq:l2}
\end{eqnarray}
Where $\mathbf{Q} =\mathbf{Q}_i \mathbf{Q}_j^T$ if $\mathbf{X}$ and $\mathbf{Y}$ are the vectors from languages $i$ and $j$.
We adapt the RCSLS loss of Eq.~(\ref{eq:rcsls}) to the unsupervised case by applying an assignment matrix to the target vectors:
\begin{eqnarray*}
\textsc{RCSLS}(\mathbf{XQ}, \mathbf{P}\mathbf{Y} ) = - 2 \text{tr~} (\mathbf{Q}^\top \mathbf{X}^\top \mathbf{PY})
   + \frac{1}{k} \left[ \sum_{\substack{\mathbf{x}\in \mathbf{X},\\ \mathbf{z} \in \mathcal{N}_{\mathbf{P}\mathbf{Y}} (\mathbf{Q}^\top \mathbf{x})}} \mathbf{z}^\top \mathbf{Q}^\top \mathbf{x} +  \sum_{\substack{\mathbf{y}\in \mathbf{PY},\\\mathbf{Q}^\top \mathbf{z} \in \mathcal{N}_X (\mathbf{y})}} \mathbf{z}^\top \mathbf{Q} \mathbf{y} \right].
\end{eqnarray*}

\paragraph{Efficient optimization.}
Directly optimizing Eq.~(\ref{eq:simple}) is computationally prohibitive since $N^2$ terms are involved.
We use a stochastic procedure where $N$ pairs $(i,j)$ of languages are sampled at each iteration and updated.
We sample pairs according to the weights $\alpha_{ij}$.
The RCSLS loss being slower to compute than the $\ell_2$ loss, we first optimize the latter for a couple of epochs before switching to the former.
The $\ell_2$ loss is optimized with the same procedure as in~\citet{grave18}:
alternate minimization with a Sinkhorn algorithm for the assignment matrix on small batches.
We then switch to a cruder optimization scheme when minimizing the RCSLS loss: we use a greedy assignment algorithm by taking the maximum per row.
We also subsample the number of elements to compute the $K$ nearest neighbors from.
We restrict each set of vectors to its first $20$k elements.
UMH runs on a CPU with $10$ threads in less than $10$ minutes for a pair of languages and in $2$ hours for $6$ languages.


\section{Related work}

\paragraph{Bilingual word embedding alignment.}
Since the work of~\citet{Mikolov13}, 
many have proposed different approaches to align word vectors with different degrees
of supervision, from fully supervised~\citep{dinu2014improving,xing2015normalized,artetxe2016learning,joulin2018loss} 
to little supervision~\citep{smith2017offline,artetxe2017learning} 
and even fully unsupervised~\citep{Zhang17b,conneau2017word,hoshen2018iterative}.
Among unsupervised approaches, some have explicitly formulated this problem
as a distribution matching: 
\citet{cao2016distribution} align the first two moments of the word vector distributions, 
assuming Gaussian distributions.
Others~\citep{zhang2017adversarial, conneau2017word} have used a Generative Adversarial Network
framework~\citep{goodfellow2014generative}.
\citet{Zhang17b} shows that an earth mover distance can be used to refine the alignment obtained
from a generative adversarial network, drawing a connection between word embedding alignment
and Optimal Transport (OT). 
\citet{artetxe2018robust} proposes a stable algorithm to tackle distant pairs of languages and low quality embeddings. 
Closer to our work, \citet{grave18} and \citet{alvarez2018gromov} have proposed robust unsupervised 
bilingual alignment methods based on OT. Our approach takes inspiration from their work and extend them to the multilingual setting.

\paragraph{Multilingual word embedding alignment.}
\citet{nakashole2017knowledge} showed that constraining coherent word alignments between triplets of nearby languages improves the quality of
induced bilingual lexicons. 
\citet{jawanpuria2018learning} recently showed similar results on any triplets of languages in the supervised case.  
As opposed to our work, these approaches are restricted to triplets of languages and use
supervision for both the lexicon and the choice of the pivot language.  
Finally, independently of this work, \citet{chen2018unsupervised} has recently extended the bilingual method of~\citet{conneau2017word} to the multilingual setting.

\paragraph{Optimal Transport.} Optimal transport
\citep{villani2003topics,santambrogio2015optimal} provides a natural topology
on shapes and discrete probability measures \citep{peyre2017computational},
that can be leveraged with fast OT problem solvers
\citep{cuturi13,altschuler2017near}. Of particular interest is the
Gromov-Wasserstein distance \citep{gromov2007metric,memoli2011gromov}. It
has been used for shape matching under its primitive form
\citep{bronstein2006generalized,memoli2007use} and under its
entropy-regularized form \citep{solomon2016entropic}. We use the latter for our intialization.

\paragraph{Hyperalignment.}
Hyperalignment, as introduced by \citet{goodall91}, is the method of
aligning several shapes onto each other with supervision.
Recently, \citet{lorbert2012kernel} extended this supervised approach to non-Euclidean distances.
We recommend \citet{gower2004procrustes} for a thorough survey of the different extensions of Procrustes and to 
\citet{edelman1998geometry} for algorithms involving orthogonal constraints. 
For unsupervised alignment of multiple shapes, \citet{huang2007unsupervised} use a pointwise entropy based method
and apply it to face alignment.


\section{Experimental Results}

\paragraph{Implementation Details.}
We use normalized fastText word vectors trained on the Wikipedia Corpus~\citep{FastText}.
We use stochastic gradient descent (SGD) to minimize the RCSLS loss.
We run the first epoch with a batch size of $500$ and then set it to $1$k.
We set the learning rate to $0.1$ for the $\ell_2$ loss and to $25$ for the RCSLS loss in the multilingual setting, and to $50$ in the bilingual setting.
For the first two iterations, we learn the assignment with a regularized Sinkhorn. Then, for efficiency, we switch to a greedy assignment, by picking the max per row of the score matrix.
We initialize with the Gromov-Wasserstein approach applied to the first $2$k vectors and a regularization parameter $\varepsilon$ of $0.5$~\citep{peyre2016gromov}.
We use the python optimal transport package.~\footnote{POT, \url{https://pot.readthedocs.io/en/stable/}}

\paragraph{Extended MUSE Benchmark.}
We evaluate on the MUSE test datasets~\citep{conneau2017word}, learning the alignments on the following $11$ languages:
Czech, Danish, Dutch, English, French, German, Italian, Polish, Portuguese, Russian and Spanish.
MUSE bilingual lexicon are mostly translations to or from English.
For missing pairs of languages (e.g., Danish-German), we use the intersection of their translation to English to build a test set.
MUSE bilingual lexicon are built with an automatic translation system.
The construction of the new bilingual lexicon is equivalent to translate with pivot.

\paragraph{Baselines.}
We consider as baselines several bilingual alignment methods that are either supervised, i.e., Orthogonal Procrustes, GeoMM~\citep{jawanpuria2018learning} and RCSLS~\citep{joulin2018loss}, or unsupervised, i.e., Adversarial~\citep{conneau2017word}, ICP~\citep{hoshen2018iterative}, Gromov-Wasserstein (GW)~\citep{alvarez2018gromov} and  Wasserstein Procrustes (``Wass Proc.'')~\citep{grave18}.
We also compare with the unusupervised multilingual method of \citet{chen2018unsupervised}.
All the unsupervised approaches, but GW, apply the refinement step (ref.) of~\citet{conneau2017word} or of~\citet{chen2018unsupervised}.

\subsection{Triplet alignment}

\begin{table}[t]
\setlength{\tabcolsep}{4.3pt}
\centering
\begin{tabular}{l cccc c cccc c cccc}
\toprule
& \multicolumn{3}{c}{Direct} & Ind. &~& \multicolumn{3}{c}{Direct} & Ind. &~& \multicolumn{3}{c}{Direct} & Ind.\\
\cmidrule{2-4}\cmidrule{7-9}\cmidrule{12-14}
& de-\underline{en} & fr-\underline{en} & de-fr & de-fr && pt-\underline{es} & fr-\underline{es} & pt-fr & pt-fr && fi-\underline{en} & hu-\underline{en} & fi-hu & fi-hu \\
\midrule
  Pairs    & \textbf{72.3} & \textbf{80.2} & 64.5  & 61.7 &&  86.5 & \textbf{81.2} & 77.2 & 72.3 && \textbf{53.4} &  \textbf{55.9} & 45.6 & 31.9  \\
  Triplet & 71.9 & \textbf{80.2} & -    & \textbf{68.3} && \textbf{86.8} & \textbf{81.2} & - & \textbf{77.9}  && 50.2 & \textbf{55.9} & - & \textbf{42.4} \\
\bottomrule
\end{tabular}
\caption{
Accuracy averaged on both directions (source$\rightarrow$target and target$\rightarrow$source) with a NN criterion on triplet alignment with direct translation (``Direct'') and indirect translation (``Ind.'').
Indirect translation uses a pivot (source$\rightarrow$pivot$\rightarrow$target). The pivot language is underlined.
We compare our approach applied to pairs (``Pairs'') of languages and triplets (``Triplets'').
}\label{tab:triplet}
\end{table}

In these experiments, we evaluate the quality of our formulation in the simple case of language triplets.
One language acts as the pivot between the two others.
We evaluate both the direct translation to and from the pivot and the indirect translation between the two other languages.
An indirect translation is obtained by first mapping the source language to the pivot, and then from the pivot to the target language.
For completeness, we report direct translation between the source and target languages.
This experiment is inspired by the setting of~\citet{nakashole2017knowledge}.

Table~\ref{tab:triplet} compares our approach trained on language pairs and triplets.
We use a NN criterion to give insights on the quality of the dot product between mapped vectors.
We test different settings: we change the pivot or the pair of languages,
consider natural in-between and distant pivot.
We also consider languages harder to align to English, such as Finnish or Hungarian.

Overall, these variations have little impact on the performance.
The direct translation to and from the pivot is not significantly impacted by the presence or absence of a third language.
More interestingly, the indirect translation of a model trained with $3$ languages often compares favorably with the direct translation from the source to the target.
In comparison, the performance of indirect translation obtained with a bilingual model dropped by $6-8\%$.
This drop reduces to a couple of percents if a CSLS criterion is used instead of a NN criterion.

\subsection{Multilingual alignment}

In this set of experiments, we evaluate the quality of joint multilingual alignment on a larger set of $11$ languages.
We look at the impact on direct and indirect alignments.

\begin{table}[t!]
\setlength{\tabcolsep}{5.7pt}
\begin{tabular}{l cc cc cc cc cc c}
\toprule
& \multicolumn{2}{c}{en-es} & \multicolumn{2}{c}{en-fr} & \multicolumn{2}{c}{en-it} & \multicolumn{2}{c}{en-de} & \multicolumn{2}{c}{en-ru} & Avg.\\
& $\rightarrow$ & $\leftarrow$ & $\rightarrow$ & $\leftarrow$ & $\rightarrow$ & $\leftarrow$ & $\rightarrow$ & $\leftarrow$ & $\rightarrow$ & $\leftarrow$ & \\
\midrule
\multicolumn{5}{l}{\small{\emph{supervised, bilingual}}}\\
Proc. &  80.9 & 82.9 & 81.0  & 82.3 & 75.3 & 77.7 & 74.3 & 72.4 & 51.2 & 64.5 & 74.3  \\
  GeoMM & 81.4 & 85.5 & 82.1 & \underline{84.1} & - & - & 74.7 & \underline{76.7} & 51.3 & \underline{67.6} & - \\
RCSLS & \underline{84.1} & \underline{86.3} & \underline{83.3} & \underline{84.1} & \underline{79.3} & \underline{81.5} & \underline{79.1} & 76.3 & \underline{57.9} & 67.2 & \underline{77.9} \\
\midrule
\multicolumn{5}{l}{\small{\emph{unsupervised, bilingual}}}\\
GW & 81.7 & 80.4 & 81.3 & 78.9 & 78.9 & 75.2 & 71.9 & 72.8 & 45.1 & 43.7 & 71.0 \\
Adv. + ref. & 81.7 &  83.3 &  82.3 & 82.1 & 77.4 & 76.1 & 74.0 & 72.2 & 44.0 & 59.1 & 73.2\\
ICP + ref. & 82.1 & 84.1 & 82.3 & 82.9 & 77.9 & 77.5 & 74.7 & 73.0 & \textbf{47.5} & 61.8 & 74.4 \\
W-Proc. + ref. & 82.8 & 84.1 & 82.6 & 82.9 & - & - & \textbf{75.4} & 73.3 & 43.7 & 59.1 & - \\
UMH bil. & \textbf{82.5} & 84.9 & \textbf{82.9} & \textbf{83.3}& \underline{\textbf{79.4}} & \textbf{79.4} & 74.8 & 73.7 & 45.3 & 62.8 & 74.9 \\
\midrule
\multicolumn{5}{l}{\small{\emph{unsupervised, multilingual}}}\\
MAT+MPSR & \bf 82.5 & 83.7 & 82.4 & 81.8 & 78.8 & 77.4 & 74.8 & 72.9 & - & - & - \\
UMH multi. & \textbf{82.4} & \textbf{85.1} & 82.7 & \textbf{83.4} & 78.1 & \textbf{79.3} & \textbf{75.5} & \textbf{74.4} & 45.8 & \textbf{64.9} & \textbf{75.2}\\
\bottomrule
\end{tabular}
\caption{
  Accuracy of supervised and unsupervised approaches on the MUSE benchmark.
  All the approaches use a CSLS criterion.
  ``ref.'' refers to the refinement method of~\citet{conneau2017word}.
  UMH does not used a refinement step.
  Multilingual (``Multi.'') UMH is trained on $11$ languages simultaneously.
The best overall accuracy is underlined, and in bold among unsupervised methods.
}\label{tab:soa}
\end{table}

\paragraph{Direct word translation.}
Table~\ref{tab:soa} shows a comparison of UMH with other unsupervised approaches on the MUSE benchmark.
This benchmark consists of $5$ translations to and from English.
In this experiment, UMH is jointly trained on $10$ languages, plus English. The results on the remaining $5$ languages are in the appendix.
We observe a slight improvement of performance of $0.3\%$ compared to bilingual UMH, which is also consistent on the remaining $5$ languages.
This improvement is not significant but shows that our approach maintains good direct word translation.

\begin{table}[t!]
\setlength{\tabcolsep}{5pt}
\begin{tabular}{l l ccccccc}
\toprule
& Latin & Germanic & Slavic & Latin-Germ. & Latin-Slavic & Germ.-Slavic  & All \\
\midrule
\multicolumn{5}{l}{\small{\emph{supervised, bilingual}}}\\
Proc. & 75.3 & 51.6 & 47.9 & 50.2 & 46.7 & 40.9 & 50.7 \\
\midrule
\multicolumn{5}{l}{\small{\emph{unsupervised, bilingual}}}\\
W-Proc.$^*$ & 74.5 & 53.2 & 44.7 & 52.2 & 44.6 & 40.2 & 50.3 \\
UMH Bil.    & 76.6 & 54.6 & 45.5 & 53.7 & 46.3 & 40.8 & 51.7 \\
\midrule
\multicolumn{5}{l}{\small{\emph{unsupervised, multilingual}}}\\
UMH Multi. & \textbf{79.0} & \textbf{58.8} & \textbf{49.8} & \textbf{57.8} & \textbf{48.8} & \textbf{45.4} & \textbf{55.3}\\
\bottomrule
\end{tabular}
\caption{
Accuracy with a NN criterion on indirect translations averaged among and across language families.
The languages are Czech, Danish, Dutch, English, French, German, Italian, Polish, Portuguese, Russian and Spanish.
UMH is either applied independently for each pairs formed with English (``Bil.'') or jointly (``Multi.'').
W-Proc.$^*$ is our implementation of \citet{grave18} with a Gomorov-Wasserstein initialization and our optimization scheme.
}\label{tab:undir}
\end{table}

\paragraph{Indirect word translation.}
Table~\ref{tab:undir} shows the performance on indirect word translation with English as a pivot language.
We consider averaged accuracies among and across language families, i.e. Latin, Germanic and Slavic.
As expected, constraining the alignments significantly improves over the bilingual baseline, by almost $4\%$.
The biggest improvement comes from Slavic languages.
The smallest improvement is between Latin languages ($+2\%$), since they are all already well aligned with English.
In general, we observe that our approach helps the most for distant languages, but the relative improvements are
similar across all languages.

\subsection{Ablation study}
In this section, we evaluate the impact of some of our design choices on the performance of UMH.
We focus in particular on the loss function, the weighting and the initialization.
We also discuss the impact of the number of languages used for training on the performance of UMH.

\paragraph{Impact of the loss function.}
Table~\ref{tab:soa} compares bilingual UMH, with state-of-the-art unsupervised bilingual approaches on the MUSE benchmark.
All the approaches use the CSLS criterion.
UMH directly learns a bilingual mapping with an approximation of the retrieval criterion.
Bilingual UMH compares favorably with previous approaches ($+0.5\%$).
In particular, the comparison with ``W-Proc.+ref'' validates our choice of the RCSLS loss for UMH.

\begin{table}[h!]
\begin{minipage}[c]{.47\linewidth}
\vspace{16pt}
\centering
\begin{tabular}{l c c}
\toprule
  & Direct & Indirect \\
\midrule
  Uniform & 65.5 & \textbf{56.9}\\
  UMH    & \textbf{69.4} & 55.3\\
\bottomrule
\end{tabular}
\end{minipage}~~~
\begin{minipage}[c]{0.47\textwidth}
\caption{
  Comparison of uniform weights and our weighting in UMH on indirect and direct translation with a NN criterion.
}\label{tab:uniform}
\end{minipage}
\end{table}

\paragraph{Impact of weights.}
Our choice of weights $\alpha_{ij}$ favors direct translation over indirect translation.
In this set of experiments, we look at the impact of this choice over simple uniform weights.
We compare UMH with uniform weights on direct and indirect word translation with a NN criterion.
It is not surprising to see that uniform weights improve the quality of indirect word translation
at the cost of poorer direct translation.
In general, we experimentally found that our weighting makes UMH more robust when scaling to larger number of languages.

\begin{table}[h!]
  \centering
\begin{tabular}{l c c}
\toprule
  & Direct & Indirect \\
\midrule
  Convex relaxation  & 77.8 & \textbf{71.5}\\
  Gromov-Wasserstein & \textbf{78.6} & 69.8\\
\bottomrule
\end{tabular}
\caption{
  Comparison of two different initializations for UMH on direct and indirect translations with a NN criterion.
  Convex relaxation refers to the initialization of~\citet{grave18}, while ours is Gromov-Wasserstein.
  We consider the $6$ languages with mutual MUSE bilingual lexicons. 
  We only learn bilingual mappings to and from English and translate with English as a pivot.
}\label{tab:init}
\end{table}

\paragraph{Impact of the initialization.}
In Sec.~\ref{sec:WP}, we introduce a novel initialization based on the Gromov-Wassertein approach.
Instead,~\citet{grave18} consider a convex relaxation of Eq.~(\ref{eq:wass-proc}) applied to centered vectors.
Table~\ref{tab:init} shows the impact of our initialization on the performance of UMH for direct and indirect bilingual alignment.
We restrict this comparison to the $6$ languages with existing mutual bilingual lexicons in MUSE, i.e., English, French, Italian, Portuguese, Spanish and German.
We consider only direct translation to and from English and the rest of the language pairs as indirect translation.
Our initialization (Gromov-Wasserstein) outperforms the convex relaxation on what it is optimized for
but this leads to a drop of performance on indirect translation.
The most probable explanation for this difference of performance is the centering of the vectors.

\begin{table}[h!]
\centering
\begin{tabular}{l c cc c c c c c c c c}
    \toprule
    & \# languages& time && en & de & fr & es & it & pt & Avg. \\
    \midrule
    MAT+MPSR & $6$ & $5$h &&  79.6 & \bf 70.5 & \bf 82.0 & \bf 82.9 & \bf 80.9 & \bf 80.1 & \bf 79.3\\
    \midrule
    UMH & $6$  & $2$h && \bf 80.7 & 69.1 & 81.0 & 82.2 & 79.8 & 78.9 & 78.6 \\
    UMH & $11$ & $5$h && 80.4 & 68.3 & 80.5 & 81.7 & 79.0 & 78.2 & 78.0 \\
    \bottomrule
\end{tabular}
\caption{
Impact of the number of languages on  UMH performance.
We report accuracy with a CSLS criterion on the $6$ common languages.
We average accuracy of translations from and to a single language.
Numbers for MAT+MPSR are from~\citet{chen2018unsupervised}.
}\label{tab:n6}
\end{table}

\paragraph{Impact of the number of languages.}
Table~\ref{tab:n6} shows the impact of the number of languages on UMH.
We train our models on $6$ and $11$ languages, and test them on the $6$ common languages as in~\citet{chen2018unsupervised}.
We use the same default hyper-parameters for all our experiments.
These $6$ languages are English, German, French, Spanish, Italian and Portuguese. They are relatively simple to align together.
Adding new and distant languages only affects the performance of UMH by less than a percent.
This shows that UMH is robust to an increasing number of languages, even when these additional languages are quite distant from the $6$ original ones.
Finally, MAT+MPSR is slightly better than UMH ($+0.7\%$) on indirect translation.
This difference is caused by our non uniform weights $\alpha_{ij}$ that seems to have a stronger impact on small number of languages.
Note that our approach is computationally more efficient, training in 2h on a CPU instead of 5h on a GPU.

\begin{minipage}[l]{.58\linewidth}
\vspace{16pt}
\paragraph{Language tree.}
We compute a minimum spanning tree on the matrix of losses given by UMH between every pairs of languages, except English.
Three clusters appear: the Latin, Germanic and Slavic families chained as Latin-Germanic-Slavic.
This qualitiative result is coherent with Table~\ref{tab:undir}. However,
the edges between languages make little sense, e.g., the edge between Spanish and Dutch.
Our alignment based solely on embeddings is too coarse to learn subtle relations between languages.
\end{minipage}~~~
\begin{minipage}[r]{0.4\textwidth}
  \centering\includegraphics[width = .9\linewidth]{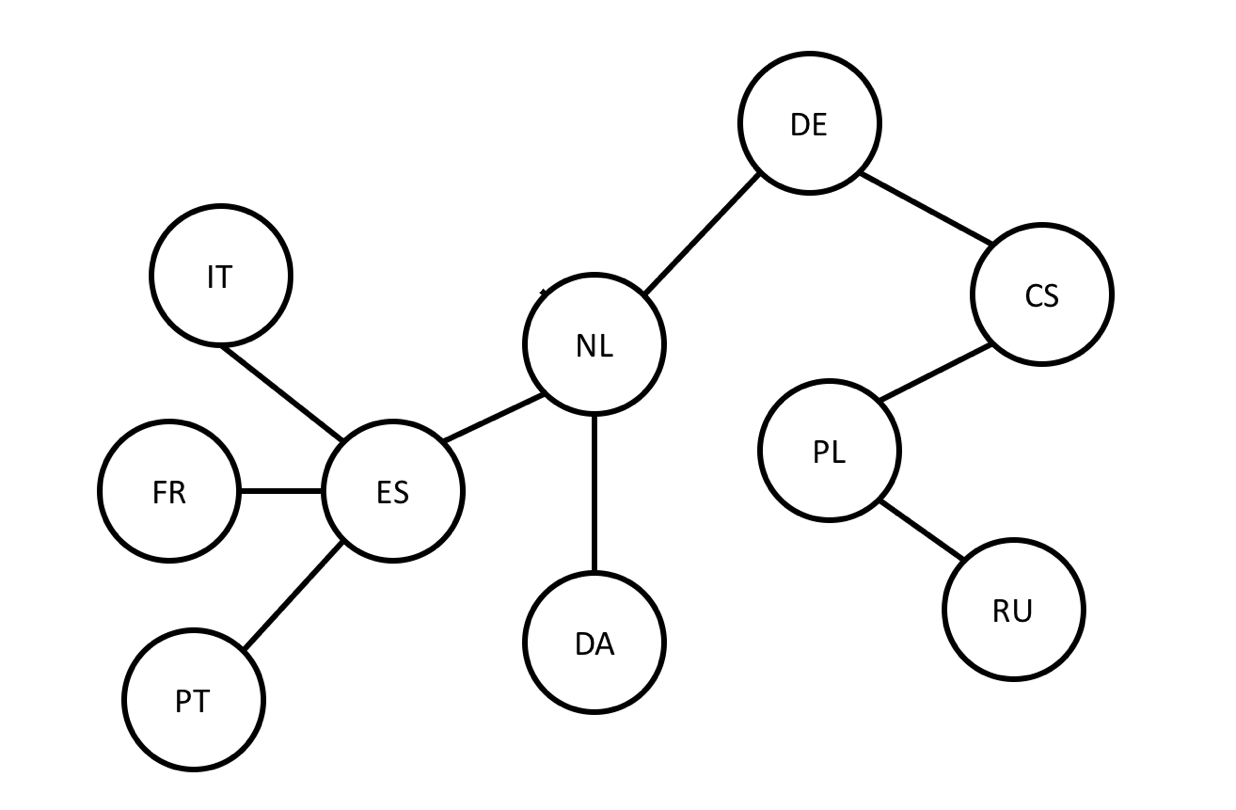}
\end{minipage}

\section{Conclusion}
This paper introduces an unsupervised multilingual alignment method that maps every language into a common space while minimizing the impact on indirect word translation.
We show that a simple extension of a bilingual formulation significantly reduces the drop of performance of indirect word translation.
Our multilingual approach also matches the performance of previously published bilingual and multilingual approaches on direct translation.
However, our current approach scales relatively well with the number of languages, but it is not clear if
such a simple approach would be enough to jointly learn the alignment of hundreds of languages.

\bibliographystyle{plainnat}      
\bibliography{biblio}

\begin{thebibliography}{41}
\providecommand{\natexlab}[1]{#1}
\providecommand{\url}[1]{\texttt{#1}}
\expandafter\ifx\csname urlstyle\endcsname\relax
  \providecommand{\doi}[1]{doi: #1}\else
  \providecommand{\doi}{doi: \begingroup \urlstyle{rm}\Url}\fi

\bibitem[Altschuler et~al.(2017)Altschuler, Weed, and
  Rigollet]{altschuler2017near}
Jason Altschuler, Jonathan Weed, and Philippe Rigollet.
\newblock Near-linear time approximation algorithms for optimal transport via
  sinkhorn iteration.
\newblock In \emph{NIPS}, 2017.

\bibitem[Alvarez-Melis and Jaakkola(2018)]{alvarez2018gromov}
David Alvarez-Melis and Tommi Jaakkola.
\newblock Gromov-wasserstein alignment of word embedding spaces.
\newblock In \emph{Proceedings of the 2018 Conference on Empirical Methods in
  Natural Language Processing}, 2018.

\bibitem[Artetxe et~al.(2016)Artetxe, Labaka, and Agirre]{artetxe2016learning}
Mikel Artetxe, Gorka Labaka, and Eneko Agirre.
\newblock Learning principled bilingual mappings of word embeddings while
  preserving monolingual invariance.
\newblock In \emph{EMNLP}, 2016.

\bibitem[Artetxe et~al.(2017)Artetxe, Labaka, and Agirre]{artetxe2017learning}
Mikel Artetxe, Gorka Labaka, and Eneko Agirre.
\newblock Learning bilingual word embeddings with (almost) no bilingual data.
\newblock In \emph{ACL}, volume~1, pages 451--462, 2017.

\bibitem[Artetxe et~al.(2018)Artetxe, Labaka, and Agirre]{artetxe2018robust}
Mikel Artetxe, Gorka Labaka, and Eneko Agirre.
\newblock A robust self-learning method for fully unsupervised cross-lingual
  mappings of word embeddings.
\newblock \emph{arXiv preprint arXiv:1805.06297}, 2018.

\bibitem[Bojanowski et~al.(2016)Bojanowski, Grave, Joulin, and
  Mikolov]{FastText}
Piotr Bojanowski, Edouard Grave, Armand Joulin, and Tomas Mikolov.
\newblock Enriching word vectors with subword information.
\newblock \emph{arXiv preprint}, arXiv:1607.04606, 2016.
\newblock URL \url{https://arxiv.org/abs/1607.04606}.

\bibitem[Bronstein et~al.(2006)Bronstein, Bronstein, and
  Kimmel]{bronstein2006generalized}
Alexander~M Bronstein, Michael~M Bronstein, and Ron Kimmel.
\newblock Generalized multidimensional scaling: a framework for
  isometry-invariant partial surface matching.
\newblock \emph{Proceedings of the National Academy of Sciences}, 103\penalty0
  (5):\penalty0 1168--1172, 2006.

\bibitem[Cao et~al.(2016)Cao, Zhao, Zhang, and Meng]{cao2016distribution}
Hailong Cao, Tiejun Zhao, Shu Zhang, and Yao Meng.
\newblock A distribution-based model to learn bilingual word embeddings.
\newblock In \emph{COLING}, 2016.

\bibitem[Chen and Cardie(2018)]{chen2018unsupervised}
Xilun Chen and Claire Cardie.
\newblock Unsupervised multilingual word embeddings.
\newblock \emph{arXiv preprint arXiv:1808.08933}, 2018.

\bibitem[Collobert et~al.(2011)Collobert, Weston, Bottou, Karlen, Kavukcuoglu,
  and Kuksa]{collobert2011natural}
Ronan Collobert, Jason Weston, L{\'e}on Bottou, Michael Karlen, Koray
  Kavukcuoglu, and Pavel Kuksa.
\newblock Natural language processing (almost) from scratch.
\newblock \emph{JMLR}, 12\penalty0 (Aug):\penalty0 2493--2537, 2011.

\bibitem[Conneau et~al.(2017)Conneau, Lample, Ranzato, Denoyer, and
  J{\'e}gou]{conneau2017word}
Alexis Conneau, Guillaume Lample, Marc'Aurelio Ranzato, Ludovic Denoyer, and
  Herv{\'e} J{\'e}gou.
\newblock Word translation without parallel data.
\newblock \emph{arXiv preprint arXiv:1710.04087}, 2017.

\bibitem[Cuturi(2013)]{cuturi13}
Marco Cuturi.
\newblock Sinkhorn distances: Lightspeed computation of optimal transport.
\newblock In \emph{NIPS}, 2013.

\bibitem[Deerwester et~al.(1990)Deerwester, Dumais, Furnas, Landauer, and
  Harshman]{deerwester1990indexing}
Scott Deerwester, Susan~T Dumais, George~W Furnas, Thomas~K Landauer, and
  Richard Harshman.
\newblock Indexing by latent semantic analysis.
\newblock \emph{Journal of the American society for information science},
  41\penalty0 (6):\penalty0 391--407, 1990.

\bibitem[Dinu et~al.(2014)Dinu, Lazaridou, and Baroni]{dinu2014improving}
Georgiana Dinu, Angeliki Lazaridou, and Marco Baroni.
\newblock Improving zero-shot learning by mitigating the hubness problem.
\newblock \emph{arXiv preprint arXiv:1412.6568}, 2014.

\bibitem[Edelman et~al.(1998)Edelman, Arias, and Smith]{edelman1998geometry}
Alan Edelman, Tom{\'a}s~A Arias, and Steven~T Smith.
\newblock The geometry of algorithms with orthogonality constraints.
\newblock \emph{SIAM journal on Matrix Analysis and Applications}, 20\penalty0
  (2):\penalty0 303--353, 1998.

\bibitem[Goodall(1991)]{goodall91}
Colin Goodall.
\newblock Procrustes methods in the statistical analysis of shape.
\newblock \emph{Journal of the Royal Statistical Society. Series B
  (Methodological)}, pages 285–339, 1991.

\bibitem[Goodfellow et~al.(2014)Goodfellow, Pouget-Abadie, Mirza, Xu,
  Warde-Farley, Ozair, Courville, and Bengio]{goodfellow2014generative}
Ian Goodfellow, Jean Pouget-Abadie, Mehdi Mirza, Bing Xu, David Warde-Farley,
  Sherjil Ozair, Aaron Courville, and Yoshua Bengio.
\newblock Generative adversarial nets.
\newblock In \emph{NIPS}, 2014.

\bibitem[Gower et~al.(2004)Gower, Dijksterhuis, et~al.]{gower2004procrustes}
John~C Gower, Garmt~B Dijksterhuis, et~al.
\newblock \emph{Procrustes problems}, volume~30.
\newblock Oxford University Press on Demand, 2004.

\bibitem[Grave et~al.(2018)Grave, Joulin, and Berthet]{grave18}
Edouard Grave, Armand Joulin, and Quentin Berthet.
\newblock Unsupervised alignment of embeddings with wasserstein procrustes.
\newblock \emph{CoRR}, abs/1805.11222, 2018.
\newblock URL \url{http://arxiv.org/abs/1805.11222}.

\bibitem[Gromov(2007)]{gromov2007metric}
Mikhail Gromov.
\newblock \emph{Metric structures for Riemannian and non-Riemannian spaces}.
\newblock Springer Science \& Business Media, 2007.

\bibitem[Hoshen and Wolf(2018)]{hoshen2018iterative}
Yedid Hoshen and Lior Wolf.
\newblock An iterative closest point method for unsupervised word translation.
\newblock \emph{arXiv preprint arXiv:1801.06126}, 2018.

\bibitem[Huang et~al.(2007)Huang, Jain, and
  Learned-Miller]{huang2007unsupervised}
Gary~B Huang, Vidit Jain, and Erik Learned-Miller.
\newblock Unsupervised joint alignment of complex images.
\newblock In \emph{ICCV}, pages 1--8. IEEE, 2007.

\bibitem[Jawanpuria et~al.(2018)Jawanpuria, Balgovind, Kunchukuttan, and
  Mishra]{jawanpuria2018learning}
Pratik Jawanpuria, Arjun Balgovind, Anoop Kunchukuttan, and Bamdev Mishra.
\newblock Learning multilingual word embeddings in latent metric space: a
  geometric approach.
\newblock \emph{arXiv preprint arXiv:1808.08773}, 2018.

\bibitem[Joulin et~al.(2018)Joulin, Bojanowski, Mikolov, J\'egou, and
  Grave]{joulin2018loss}
Armand Joulin, Piotr Bojanowski, Tomas Mikolov, Herv\'e J\'egou, and Edouard
  Grave.
\newblock Loss in translation: Learning bilingual word mapping with a retrieval
  criterion.
\newblock In \emph{EMNLP}, 2018.

\bibitem[Lorbert and Ramadge(2012)]{lorbert2012kernel}
Alexander Lorbert and Peter~J Ramadge.
\newblock Kernel hyperalignment.
\newblock In \emph{NIPS}, 2012.

\bibitem[M{\'e}moli(2007)]{memoli2007use}
Facundo M{\'e}moli.
\newblock On the use of gromov-hausdorff distances for shape comparison.
\newblock 2007.

\bibitem[M{\'e}moli(2011)]{memoli2011gromov}
Facundo M{\'e}moli.
\newblock Gromov--wasserstein distances and the metric approach to object
  matching.
\newblock \emph{Foundations of computational mathematics}, 11\penalty0
  (4):\penalty0 417--487, 2011.

\bibitem[Mikolov et~al.(2013{\natexlab{a}})Mikolov, Le, and
  Sutskever]{Mikolov13}
Tomas Mikolov, Quoc~V Le, and Ilya Sutskever.
\newblock Exploiting similarities among languages for machine translation.
\newblock \emph{arXiv preprint}, arXiv:1309.4168v, 2013{\natexlab{a}}.
\newblock URL \url{https://arxiv.org/abs/1309.4168}.

\bibitem[Mikolov et~al.(2013{\natexlab{b}})Mikolov, Sutskever, Chen, Corrado,
  and Dean]{mikolov2013distributed}
Tomas Mikolov, Ilya Sutskever, Kai Chen, Greg~S Corrado, and Jeff Dean.
\newblock Distributed representations of words and phrases and their
  compositionality.
\newblock In \emph{NIPS}, 2013{\natexlab{b}}.

\bibitem[Nakashole and Flauger(2017)]{nakashole2017knowledge}
Ndapandula Nakashole and Raphael Flauger.
\newblock Knowledge distillation for bilingual dictionary induction.
\newblock In \emph{Proceedings of the 2017 Conference on Empirical Methods in
  Natural Language Processing}, pages 2497--2506, 2017.

\bibitem[Peyr{\'e} et~al.(2016)Peyr{\'e}, Cuturi, and Solomon]{peyre2016gromov}
Gabriel Peyr{\'e}, Marco Cuturi, and Justin Solomon.
\newblock Gromov-wasserstein averaging of kernel and distance matrices.
\newblock In \emph{ICML}, 2016.

\bibitem[Peyr{\'e} et~al.(2017)Peyr{\'e}, Cuturi,
  et~al.]{peyre2017computational}
Gabriel Peyr{\'e}, Marco Cuturi, et~al.
\newblock Computational optimal transport.
\newblock Technical report, 2017.

\bibitem[Santambrogio(2015)]{santambrogio2015optimal}
Filippo Santambrogio.
\newblock Optimal transport for applied mathematicians.
\newblock \emph{Birk{\"a}user, NY}, pages 99--102, 2015.

\bibitem[Schönemann(1966)]{schönemann66}
Peter~H. Schönemann.
\newblock A generalized solution of the orthogonal procrustes problem.
\newblock \emph{Psychometrika}, 31(1):1–10, 1966.

\bibitem[Smith et~al.(2017)Smith, Turban, Hamblin, and
  Hammerla]{smith2017offline}
Samuel~L Smith, David~HP Turban, Steven Hamblin, and Nils~Y Hammerla.
\newblock Offline bilingual word vectors, orthogonal transformations and the
  inverted softmax.
\newblock \emph{arXiv preprint arXiv:1702.03859}, 2017.

\bibitem[Solomon et~al.(2016)Solomon, Peyr{\'e}, Kim, and
  Sra]{solomon2016entropic}
Justin Solomon, Gabriel Peyr{\'e}, Vladimir~G Kim, and Suvrit Sra.
\newblock Entropic metric alignment for correspondence problems.
\newblock \emph{ACM Transactions on Graphics (TOG)}, 35\penalty0 (4):\penalty0
  72, 2016.

\bibitem[Villani(2003)]{villani2003topics}
C{\'e}dric Villani.
\newblock \emph{Topics in optimal transportation}.
\newblock Number~58. American Mathematical Soc., 2003.

\bibitem[Xing et~al.(2015{\natexlab{a}})Xing, Wang, Liu, and Lin]{Xing15}
Chao Xing, Dong Wang, Chao Liu, and Yiye Lin.
\newblock Normalized word embedding and orthogonal trans- form for bilingual
  word translation.
\newblock In \emph{NAACL}, 2015{\natexlab{a}}.

\bibitem[Xing et~al.(2015{\natexlab{b}})Xing, Wang, Liu, and
  Lin]{xing2015normalized}
Chao Xing, Dong Wang, Chao Liu, and Yiye Lin.
\newblock Normalized word embedding and orthogonal transform for bilingual word
  translation.
\newblock In \emph{NAACL}, pages 1006--1011, 2015{\natexlab{b}}.

\bibitem[Zhang et~al.(2017{\natexlab{a}})Zhang, Liu, Luan, and Sun]{Zhang17b}
Meng Zhang, Yang Liu, Huanbo Luan, and Maosong Sun.
\newblock Earth mover’s distance minimization for unsupervised bilingual
  lexicon induction.
\newblock In \emph{EMNLP}, 2017{\natexlab{a}}.

\bibitem[Zhang et~al.(2017{\natexlab{b}})Zhang, Liu, Luan, and
  Sun]{zhang2017adversarial}
Meng Zhang, Yang Liu, Huanbo Luan, and Maosong Sun.
\newblock Adversarial training for unsupervised bilingual lexicon induction.
\newblock In \emph{ACL}, volume~1, pages 1959--1970, 2017{\natexlab{b}}.

\end{thebibliography}

\newpage
\section*{Appendix}
We present detailed results of our experiments on $6$ and $11$ languages.

\begin{table}[h]
  \centering
\setlength{\tabcolsep}{5.7pt}
\begin{tabular}{l cccccccccc}
\toprule
& en-fr & en-es & en-it & en-pt & en-de & en-da & en-nl & en-pl & en-ru & en-cs \\
\midrule
\multicolumn{5}{l}{\small{\emph{unsupervised, unconstrained}}}\\
Bil. - NN   & 80.1 & 80.9 & 76.5 & 77.7 & 73.1 & 55.7 & 72.9 & 55.9 & 43.7 & 49.3 \\
Bil. - CSLS & 82.9 & 82.5 & 79.4 & 81.7 & 74.8 & 61.7 & 76.7 & 57.7 & 45.3 & 52.6 \\
\midrule
\multicolumn{5}{l}{\small{\emph{unsupervised, constrained}}}\\
Mul. - NN   & 79.7 & 81.3 & 76.2 & 78.1 & 73.3 & 55.8 & 71.6 & 54.2 & 44.1 & 50.9 \\
Mul. - CSLS & 82.7 & 82.4 & 78.1 & 81.3 & 75.5 & 60.4 & 76.3 & 56.1 & 45.8 & 53.5 \\
\midrule
& fr-en & es-en & it-en & pt-en & de-en & pl-en & ru-en & da-en & nl-en & cs-en \\
\midrule
\multicolumn{5}{l}{\small{\emph{unsupervised, unconstrained}}}\\
Bil. - NN   & 80.3 & 82.5 & 77.9 & 79.7 & 71.4 & 63.9 & 74.2 & 67.7 & 60.7 & 61.3 \\
Bil. - CSLS & 83.3 & 84.9 & 79.4 & 82.2 & 73.7 & 65.9 & 77.1 & 67.7 & 62.8 & 62.5 \\
\midrule
\multicolumn{5}{l}{\small{\emph{unsupervised, constrained}}}\\
Mul. - NN  & 80.6 & 81.8 & 77.3 & 78.8 & 72.1 & 64.6 & 73.1 & 68.3 & 62.3 & 63.9 \\
Mul. - CSLS& 83.4 & 85.1 & 79.3 & 81.4 & 74.4 & 67.0 & 76.2 & 68.9 & 64.9 & 64.4 \\
\bottomrule
\end{tabular}
\caption{Full results on direct translation with the $11$ languages and both NN and CSLS criteria for the UMH method.
Bil. stands for bilingual, and Mul. stands for multilingual.
}
\end{table}

\begin{table}[h]
  \centering
\setlength{\tabcolsep}{5.7pt}
\begin{tabular}{l cccccccccc}
\toprule
$\rightarrow$ & fr & es & it & pt & de & da & nl & pl & ru & cs \\
\midrule
  fr &   -  & 82.5 & 82.1 & 77.1 & 69.3 & 53.1 & 67.8 & 47.1 & 40.7 & 42.7 \\
  es & 84.9 &   -  & 83.3 & 86.5 & 68.3 & 54.7 & 66.6 & 48.6 & 44.2 & 46.7 \\
  it & 86.5 & 86.3 &   -  & 79.8 & 66.6 & 51.8 & 66.0 & 50.8 & 39.4 & 43.9\\
  pt & 82.9 & 91.5 & 79.9 &   -  & 63.0 & 52.2 & 63.7 & 48.8 & 39.7 & 44.4 \\
  de & 73.0 & 66.4 & 68.5 & 58.6 &   -  & 59.5 & 69.8 & 48.6 & 39.8 & 44.9\\ 
  da & 59.1 & 63.4 & 59.1 & 61.0 & 65.4 &   -  & 65.4 & 44.5 & 34.0 & 42.2\\
  nl & 69.0 & 70.1 & 68.6 & 67.7 & 75.9 & 58.8 &   -  & 47.7 & 40.4 & 44.9\\
  pl & 62.5 & 66.1 & 61.4 & 63.3 & 62.3 & 49.3 & 59.9 &   -  & 53.3 & 57.7\\ 
  ru & 60.1 & 61.4 & 57.1 & 57.3 & 55.9 & 46.0 & 54.2 & 56.3 &   -  & 52.2\\
  cs & 60.7 & 62.9 & 59.4 & 61.4 & 59.6 & 53.6 & 58.5 & 59.7 & 49.2 &   - \\
\bottomrule
\end{tabular}
\caption{Full results on indirect translation with the $11$ languages with a CSLS criterion.
}
\end{table}

\begin{table}[h]
  \centering
\setlength{\tabcolsep}{5.7pt}
\begin{tabular}{l cccccc}
\toprule
$\rightarrow$  & en & fr & es & it & pt & de \\
\midrule
  en &  -  & 82.7 & 82.5 & 78.9 & 82.0 & 75.1 \\
  fr &83.1 &  -   & 82.7 & 82.5 & 77.5 & 69.8 \\
  es &85.3 & 85.1 &  -   & 83.3 & 86.3 & 68.7 \\
  it &79.9 & 86.7 & 87.0 &  -   & 80.4 & 67.5 \\
  pt &82.1 & 83.6 & 91.7 & 81.1 &  -   & 64.4 \\
  de &75.5 & 73.5 & 67.2 & 68.7 & 59.0 &  -   \\
\bottomrule
\end{tabular}
\caption{Full results of our model trained on $6$ languages with a CSLS criterion.
}
\end{table}

\end{document}